\documentclass[journal]{IEEEtran}

\usepackage[T1]{fontenc}% optional T1 font encoding

\usepackage{cite}

\usepackage{multirow}
\usepackage{graphicx}
\usepackage{amssymb}
\usepackage{amsmath}
\usepackage{booktabs}
\usepackage{array}
\newcolumntype{L}[1]{>{\raggedright\let\newline\\\arraybackslash\hspace{0pt}}m{#1}}
\newcolumntype{C}[1]{>{\centering\let\newline\\\arraybackslash\hspace{0pt}}m{#1}}
\newcolumntype{R}[1]{>{\raggedleft\let\newline\\\arraybackslash\hspace{0pt}}m{#1}}

\usepackage{lipsum}
\usepackage{changepage}
\usepackage{color,soul}
\usepackage{enumitem}   

\usepackage{pifont}

\usepackage{tikz}

\usepackage{lipsum}

\let\OLDthebibliography\thebibliography
\renewcommand\thebibliography[1]{
  \OLDthebibliography{#1}
  \setlength{\parskip}{0pt}
  \setlength{\itemsep}{0pt plus 0.3ex}
}

\usepackage{array}

\makeatletter
\newcommand{\thickhline}{%
    \noalign {\ifnum 0=`}\fi \hrule height 1pt
    \futurelet \reserved@a \@xhline
}
\newcolumntype{"}{@{\hskip\tabcolsep\vrule width 1pt\hskip\tabcolsep}}
\makeatother

\usepackage[ruled,vlined]{algorithm2e}

\usepackage{framed,multirow}

\usepackage{latexsym}

\usepackage{url}
\usepackage{xcolor}

\usepackage{hyperref}

\usepackage{bm}

\begin{document}

\title{Diffusion Tensor Estimation with Uncertainty Calibration} % \thanks{Supported by organization x.}}

\author{Davood Karimi$^*$, Simon K. Warfield, and Ali Gholipour \\ Computational Radiology Laboratory, Boston Children's Hospital, Harvard Medical School \\ Boston, MA 02115, USA \\ $^*$davood.karimi@childrens.harvard.edu}

%\thanks{Relevant code can be found at: \url{https://github.com/bchimagine/calibrated_dti}. } }

\maketitle

\begin{abstract}

It is highly desirable to know how uncertain a model's predictions are, especially for models that are complex and hard to understand as in deep learning. Although there has been a growing interest in using deep learning methods in diffusion-weighted MRI, prior works have not addressed the issue of model uncertainty. Here, we propose a deep learning method to estimate the diffusion tensor and compute the estimation uncertainty. Data-dependent uncertainty is computed directly by the network and learned via loss attenuation. Model uncertainty is computed using Monte Carlo dropout. We also propose a new method for evaluating the quality of predicted uncertainties. We compare the new method with the standard least-squares tensor estimation and bootstrap-based uncertainty computation techniques. Our experiments show that when the number of measurements is small the deep learning method is more accurate and its uncertainty predictions are better calibrated than the standard methods. We show that the estimation uncertainties computed by the new method can highlight the model's biases, detect domain shift, and reflect the strength of noise in the measurements. Our study shows the importance and practical value of modeling prediction uncertainties in deep learning-based diffusion MRI analysis.

\end{abstract}


\section{Introduction}
\label{Introduction_section}

\subsection{Diffusion tensor imaging}

Diffusion tensor imaging (DTI) is one of the most common models in diffusion weighted magnetic resonance imaging (dMRI). In DTI, the diffusion signal is modeled as $S(q,b)/S_0= \exp \big(- b q^T D q   \big)$, where $q$ and $b$ denote, respectively, the direction and strength of the applied gradient, $S_0$ is the baseline signal, and $D$ is the 3-by-3 diffusion tensor \cite{basser2011microstructural}. An eigen-decomposition of $D$ can be used to obtain the direction of the strongest diffusion as well as parameters such as mean diffusivity (MD) and fractional anisotropy (FA). Although there exist more complex and more accurate models of dMRI signal \cite{assaf2005composite,zhang2012noddi}, because of its lower acquisition requirements and simpler interpretation of the resulting biomarkers DTI is widely used to study various neurodevelopmental factors and neurological disorders (e.g.,  \cite{kubicki2007review,barnea2004white}).

After log transformation of the normalized signal $S/S_0$, a linear system of equations can be formed for estimating the diffusion tensor. Different  least squares-based DTI estimation approaches have been proposed \cite{koay2006unifying}. Recently, deep learning (DL) methods have been proposed to estimate the diffusion tensor \cite{tian2020deepdti,li2021superdti} and many other dMRI parameters \cite{golkov2016,nath2019deep}. The use of DL in dMRI has been promoted and justified based on the ability of DL models to learn spatial correlations, their ability to learn highly complex regression functions, and availability of large high-quality dMRI datasets that can be used to train these models. Many of these works have claimed a significant improvement in estimation accuracy and/or reduction in the required number of measurements. However, important issues such as estimation uncertainty and generalizability to different dMRI acquisition settings has been neglected by most prior studies. With regard to uncertainty estimation, we are aware of only two studies where Lasso bootstrap and ensemble methods were used to compute uncertainty in estimating the NODDI parameters \cite{ye2020improved,qin2021super}.

Estimation errors in DTI may arise due to such factors as measurement noise, inadequacy of the number of measurements, or the poor choice of directions of diffusion-sensitizing gradients. Furthermore, like any other mathematical model, DTI is unable to fully represent the underlying biophysics. The DTI model wrongly assumes a single Gaussian diffusion process, and it fails to account for crossing fibers \cite{assaf2014inferring,seunarine2014multiple}. Therefore, regardless of the measurement protocol and model fitting approach, estimation errors are unavoidable. The ability to predict the anticipated estimation errors or estimation uncertainty can be useful in interpreting and further processing the results. For example, uncertainty in the DTI-derived orientation of the major fiber (i.e., major eigenvector of the tensor) can be useful in probabilistic tractography. Likewise, DTI-derived biomarkers such as FA and MD can be more accurately interpreted and compared across subjects if they come with a well-calibrated measure of estimation uncertainty.

\subsection{Bootstrap methods}

The utility of proper uncertainty estimation in DTI has long been recognized. For classical least squares-based DTI estimation methods, analytical perturbation techniques and empirical bootstrap techniques have been used to compute the estimation uncertainty, with bootstrap techniques being much more common. Jones, \cite{jones2003determining}, used a bootstrap method to compute a cone of uncertainty for the orientation of the major eigenvector of the diffusion tensor. From two complete dMRI volumes acquired with identical gradient tables, he selected $n$ bootstrap sets to obtain $n$ tensor estimations. For each voxel, he then computed a mean dyadic tensor from these $n$ estimations. He computed the angle between the orientation of the major eigenvector of each of the $n$ tensors and that of the mean tensor, and proposed the 95 percentile of these angles as a measure of estimation uncertainty. A similar approach has been followed in many studies to also compute the estimation uncertainty for DTI-derived parameters such as FA and MD, where uncertainty is defined as the standard deviation of bootstrap-estimated values \cite{landman2007effects,zhu2008optimized}. While earlier works used a repetition bootstrap approach that required multiple measurements along the same diffusion-sensitizing gradient, later studies proposed wild bootstrap and residual bootstrap methods that did not require multiple measurements with the same diffusion-sensitizing gradient and that were compatible with noise heteroscedasticity in dMRI \cite{whitcher2008using,zhu2008optimized}.

Prior bootstrap-based studies have evaluated the estimation uncertainty in a global and often qualitative and subjective manner. For example, they have pointed out that the estimation uncertainty is lower in the location of major coherent white matter tracts and for simulated tensors with high anisotropy, and that it is higher in voxels where more than one white matter tracts intersect \cite{jones2003determining,yuan2008note}. Based on similar global and/or qualitative assessments, some works have questioned the validity of estimation uncertainty with bootstrap methods in some scenarios such as oblate tensors and voxels with crossing fibers \cite{yuan2008note}. Moreover, most of the above works relied on simulation experiments and used Monte Carlo methods as the gold standard for comparison \cite{zhu2008optimized,chung2006comparison,zhu2009evaluation}.

\subsection{Uncertainty in deep learning-based regression}

\subsubsection{\textbf{Definitions}}

Consider a regression problem where the goal is to find a data generating function $f$ from a set of noisy observations $y_i \in {\rm I\!R}$ and corresponding input $x_i \in {\rm I\!R}^d$:

\begin{equation}  \label{eq:regression}
\footnotesize
y_i= f(x_i) + \epsilon_i.
\end{equation}

Here, $\epsilon$ is the irreducible noise, which can be due to the inherently stochastic nature of the problem or omission of certain explanatory variables from $x$. Assuming $\epsilon$ has a mean of zero, the goal is to find $\hat{f}$ that is as close as possible to $f$. Assuming that the uncertainty in $\hat{f}$ and the uncertainty due to noise are independent, we can write the total variance as $\sigma^2= \sigma_{\text{model}}^2 + \sigma_{\text{noise}}^2$ \cite{pearce2018high,sluijterman2021evaluate}. Here, $\sigma_{\text{model}}$ is due to misspecification of $\hat{f}$, which can be either because $f$ is outside the hypothesis space of $\hat{f}$ or due to parameter misspecification. Since deep neural networks can represent any function of practical interest \cite{hornik1989multilayer}, $\sigma_{\text{model}}^2$ is assumed to be due to misspecification of network weights and is usually referred to as model uncertainty or epistemic uncertainty. $\sigma_{\text{noise}}^2$, on the other hand, is referred to as noise variance, irreducible variance, or aleatoric uncertainty. Some studies have considered these two terms separately, whereas many other studies consider the overall uncertainty term $\sigma^2$, which they usually refer to as predictive uncertainty. In machine learning, uncertainty may also arise due to a mismatch between the training and test data distributions. 

Given a test data sample, $x_i$, many models predict a point estimate $\hat{y}_i$. In uncertainty estimation, the goal is to also estimate $\sigma$, or the distribution of $y_i$. Calibration, also known as reliability, refers to the requirement that the predicted probabilities should be consistent with the frequency of observations.

\subsubsection{\textbf{Prediction uncertainty and calibration}}

Although DL models are rightly criticized for being too complex and hard to interpret, they are increasingly used to address safety-critical applications. Therefore, proper assessment of reliability and uncertainty of these models has high practical utility \cite{abdar2021review}. Yet, it has been shown that DL models produce overconfident and poorly calibrated predictions \cite{guo2017,lakshminarayanan2017}. Most prior studies have addressed classification problems \cite{abdar2021review}.
Uncertainty estimation of DL-based regression has received less attention. In fact, while for classification there is a widely-accepted definition of calibration, for regression different definitions have been proposed. A classifier that assigns label $\hat{y}_i$ to the data sample $x_i$ with probability $p_i$ is said to be calibrated if $E[\hat{y}_i=y_{\text{true}}]= p_i$. Kuleshov et al. proposed a natural extension of this definition to regression problems \cite{kuleshov2018accurate}. Consider a regression function that maps each input $x_i$ to a probability distribution $p_i(y)$. Denoting the cumulative distribution function (CDF) of $p_i$ with $F_i$, they suggest that the regression function is perfectly calibrated if:

\begin{equation}  \label{eq:Kuleshov}
\footnotesize
\frac{\sum_{i=1}^N \mathbb{I} \{ y_i < F_i^{-1}(p) \} }{N} \xrightarrow{ \; \; N \rightarrow \infty  \; \; } p 
\end{equation}

This definition is similar to the definitions based on credible intervals that have been used in many prior works \cite{gneiting2007probabilistic}. However, it has been suggested that this definition of calibration should be referred to as average calibration because it reflects the average calibration over the entire dataset rather than calibration on each individual data point \cite{song2019distribution,chung2021uncertainty}. It is possible to have uninformative models with perfect average calibration. To avoid this problem, some studies have proposed to also quantify the sharpness/concentration of the predicted distributions \cite{chung2021uncertainty}. Sharpness is usually quantified as the standard deviation of the predicted uncertainty or confidence intervals, and it is independent of the true distribution. An alternative definition that aims to overcome this shortcoming of Eq. \ref{eq:Kuleshov} is adversarial group calibration \cite{zhao2020individual}, which essentially requires calibration over any random subset of the input space. Moreover, it has been shown that using the definition in Eq. \ref{eq:Kuleshov} one can calibrate any output distribution, even distributions that are uncorrelated with the empirical uncertainty \cite{levi2019evaluating}. Levi et al. have proposed an alternative definition that avoids the averaging. Denoting the predicted mean and variance, respectively, with $\mu(x)$ and $\sigma(x)$, perfect calibration is defined as one that follows:

\begin{equation}  \label{eq:Levi}
\footnotesize
E_{x,y} \{ (\mu(x)-y)^2|\sigma(x)^2=\sigma^2 \} = \sigma^2
\end{equation}

The idea behind this definition is that uncertainty, quantified as standard deviation $\sigma$, should match the expected error. This definition, has similarities with definitions that have been proposed for classification \cite{guo2017,naeini2015}.

Another set of metrics used to assess uncertainty calibration are based on proper scoring rules such as the logarithmic score and interval score \cite{gneiting2007strictly}. Alternatively, a number of works have followed distribution-free approaches to defining uncertainty and evaluating calibration. The central concept in these methods is to estimate prediction intervals that capture most of the observations while being as narrow as possible \cite{sluijterman2021evaluate,khosravi2010lower}. This has sometimes been referred to as the high-quality principle \cite{pearce2018high}. Two widely used measures in these methods are Prediction Interval Coverage Probability (PICP) and Mean Prediction Interval Width (MPIW). Let us denote the lower and upper bounds of the prediction interval for the $i^{\text{th}}$ data point with ${y_L}_i$ and ${y_U}_i$. Then, PICP and MPIW are defined as \cite{pearce2018high,khosravi2010lower}:

\begin{equation}  \label{eq:PICP_MPIW}
\footnotesize
\text{PICP} = \frac{1}{n} \sum_{i=1}^n \mathbb{I} \{ {y_L}_i \leq y_i \leq {y_U}_i   \} ; \; \text{MPIW} = \frac{1}{n} \sum_{i=1}^n {y_U}_i - {y_L}_i
\end{equation}

A common objective is to minimize MPIW subject to $\text{PICP} \leq (1-\alpha)$, where $\alpha$ is typically 0.01 or 0.05. 

Although there is much overlap between some of the criteria that have been discussed above, not all these criteria always agree \cite{gneiting2007strictly,chung2020beyond}. A combination of several of these measures may be necessary to produce a complete picture of estimation uncertainty in a specific application \cite{chung2021uncertainty}.

\subsubsection{\textbf{Methods of estimating uncertainty}}

Bayesian neural networks present a natural framework for computing model uncertainty \cite{neal2012bayesian}. However, they are difficult to implement and computationally demanding. Moreover, the quality of their uncertainty predictions depends on the correctness of the prior distribution and the degree of approximation. A practical method based on a Bayesian inference interpretation of Monte Carlo dropout was proposed in \cite{gal2016}. Another interpretation of dropout is in terms of model ensembles. Based on this interpretation, it has been suggested that model ensembles present an effective approach to quantifying model uncertainty \cite{lakshminarayanan2017}. Early studies used bootstrap methods to train ensembles of neural networks for uncertainty estimation \cite{heskes1997practical}. Recent studies have shown that random initialization of the network weights is sufficient to train an ensemble of models that are diverse enough to produce good uncertainty estimates. Uncertainty is computed as the standard deviation of the predictions of the models \cite{lakshminarayanan2017}.

\subsubsection{\textbf{Calibration techniques}}

Some studies have aimed at training well-calibrated models. Quantile regression methods and methods for directly estimating the cumulative distribution function (CDF) have been developed \cite{koenker2001quantile,tagasovska2018single}. Another work proposed training calibrated regression methods based on the concept of learned loss attenuation \cite{gal2016}. This approach amounts to down-weighing the less certain samples in the prediction loss term at the cost of an uncertainty penalty term. Several studies have proposed training loss functions based on PCIP and MPIW. Early loss functions were not differentiable \cite{khosravi2010lower} and were optimized using such methods as genetic algorithms and  particle swarm optimization. Recently, loss functions that can be minimized with gradient descent have been proposed \cite{pearce2018high}.

Post-hoc calibration methods work on models that have already been trained. They learn a function that transforms poorly-calibrated uncertainties produced by a trained model such that they are better calibrated. This transformation is usually learned on a calibration dataset. Different forms of transformation have been propsoed \cite{kuleshov2018accurate,guo2017}. Quantile-based calibration is common. However, recent works have propsoed more general distribution-based calibration using Gaussian processes \cite{song2019distribution} and  maximum mean discrepancy distribution matching \cite{cui2020calibrated}. On the other hand, many studies have used a simple scaling to re-calibrate the uncertainty estimations of a trained model. This is usually called scaling, temperature scaling, or Platt scaling.

\subsection{Contributions of this work}

To the best of our knowledge, no prior work has assessed bootstrap-computed uncertainties in terms of standard measures of calibration. Moreover, prior DL-based works on DTI estimation have ignored the issue of estimation uncertainty. Estimation uncertainty is especially relevant for DL techniques since they have a very large number of parameters and involve multi-stage data processing architectures that are difficult or impossible to disentangle and understand.

In this paper we aim to address these gaps in research. Specifically:

\begin{itemize}

\item We propose methods for computing well-calibrated uncertainties in DL-based DTI estimation.

\item We propose simple visual and quantitative methods for evaluating and comparing different uncertainty estimates.

\item We compare standard least-squares and bootstrap methods with the propsoed DL-based techniques in terms of estimation accuracy and uncertainty calibration.

\item We show how the computed uncertainties can be used to detect the biases, shortcomings, and failures of DL-based DTI estimation.

\end{itemize}

\section{Materials and methods}

\subsection{Proposed DTI estimation method}

\begin{figure*}[!htb]
  \centering
  \centerline{\includegraphics[width=18cm]{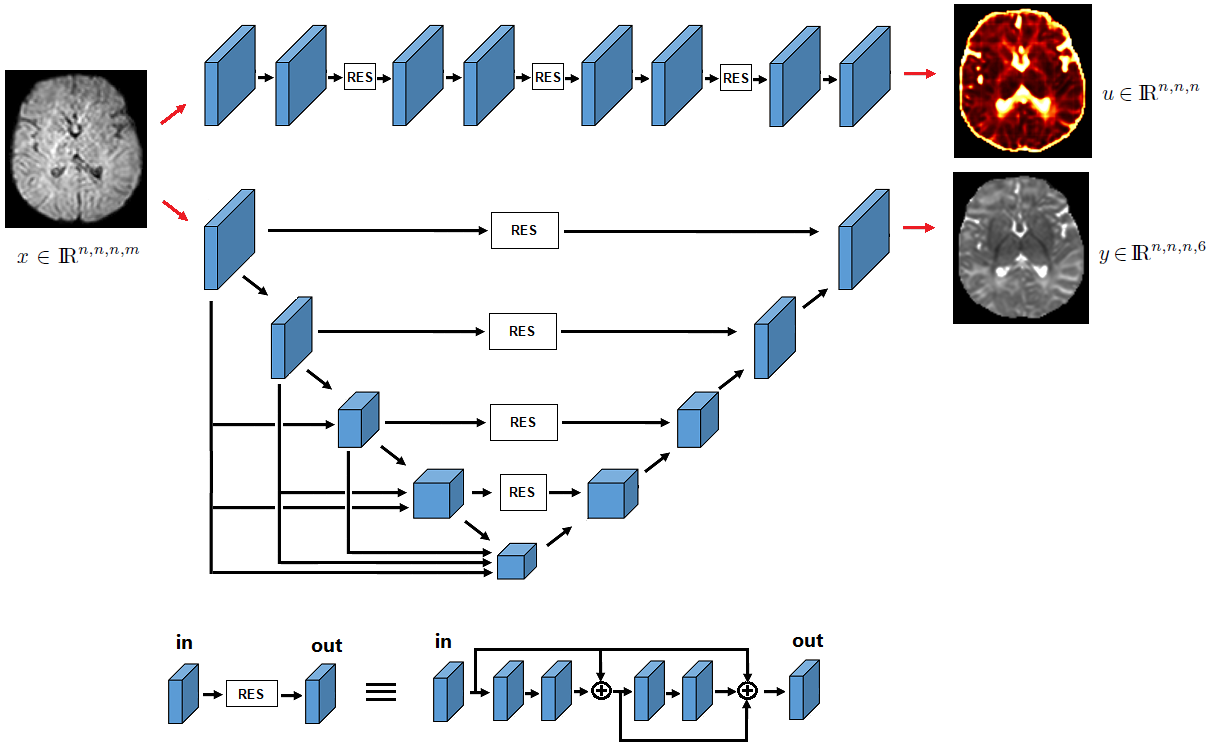}}
\caption{\footnotesize{The proposed network to compute the diffusion tensor and estimation uncertainty. The residual block, denoted as RES, is shown at the bottom of the figure. ($n$ is the patch size and $m$ is the number of measurements.)}}
\label{fig:network}
\end{figure*}

We propose a two-branch deep neural network to compute the diffusion tensor and estimation uncertainty in a single framework (Figure \ref{fig:network}). The main branch is a fully convolutional network (FCN) with an encoder-decoder architecture. Compared with a standard UNet, this architecture has additional dense connections between different stages of the encoder section. The exact network architecture is not the focus of this study. The main point is that for accurate tensor estimation we would like a model that can learn spatial correlations. Our experiments have shown that a model that works on a single voxel or a very small patch has significantly lower estimation accuracy than a model that works on large input patches. Our network takes $48^3$-voxel patches as input and computes the diffusion tensor for all voxels in that patch. The uncertainty estimation branch, on the other hand, consists of a series of convolutions with a kernel size of one. In other words, this branch only uses 1D convolutions. This is to ensure that the uncertainty estimated for each voxel depends on the signal in that voxel alone, because we want this branch to estimate data-dependent/aleatoric uncertainty. The network details are shown in Figure \ref{fig:network}. The idea of training a neural network to estimate both a signal mean and signal variance is old \cite{nix1994estimating}. Recent DL works have also used similar techniques for classification and regression applications \cite{devries2018learning,lakshminarayanan2017}. However, those works have employed a single network backbone with a final network section that splits into two heads to estimate the target variable and uncertainty. Our model, on the other hand, has two separate branches as justified above. 

Given an input dMRI patch $x \in {\rm I\!R}^{n,n,n,m}$, the network predicts the tensor and voxel-wise estimation uncertainty, respectively $y \in {\rm I\!R}^{n,n,n,6}$ and $u \in {\rm I\!R}^{n,n,n}$, for that patch. We train the network to jointly estimate both the tensor values and the estimation uncertainty using the following loss function:

\begin{equation}  \label{eq:our_loss}
\footnotesize
\mathcal{L}(\theta) = \mathlarger{\mathlarger{\sum}}_i \Bigg\{\mathlarger{\mathlarger{\sum}}_j \bigg( \frac{ | y_i^j - {y_g}_i^j |  }{ \exp (u_i) } \bigg) + \lambda u_i   \Bigg\}
\end{equation}

In this loss function, indices $i$ and $j$ refer to voxels and tensor elements, respectively. Moreover, $\theta$ denotes the network parameters, $y_g$ denotes the ground truth tensor, and $\lambda$ is the regularization constant, which was set to empirically-selected value of 1 in all our experiments. This loss function is similar to both the learned loss attenuation \cite{lakshminarayanan2017,gal2016} and prediction interval-based loss functions \cite{pearce2018high} proposed in prior works. The intuition behind this loss function is to allow the model to incur a smaller estimation error loss for less certain voxels at the cost of an uncertainty penalty. A similar loss can be derived based on data log-likelihood, assuming a Gaussian noise distribution \cite{gal2016,lakshminarayanan2017}. Compared with the loss functions based on prediction intervals \cite{pearce2018high,khosravi2010lower}, the first and the second terms in Eq. \eqref{eq:our_loss} are related to PCIP and MPIW, respectively.

Note that $u$ estimated above can only represent the data-dependent (aleatoric) uncertainty. To compute the model/epistemic uncertainty, we use Monte-Carlo dropout \cite{gal2016,sluijterman2021evaluate}. We train the main branch of the network using a dropout rate of 0.90 (chosen empirically) in all layers. On a test dMRI volume, we compute 100 different tensor values by generating random dropout masks. From these 100 tensor estimates, we compute the cone of uncertainty as the 95 percentile of the variation in the orientation of the major tensor eigenvector \cite{jones2003determining}, which we denote as $\theta_{95}^{\text{DL}}$. As commonly done in bootstrap methods \cite{tibshirani1993introduction} as well as in prior DTI estimation studies \cite{zhu2008optimized,zhu2009evaluation}, we compute the standard deviation in the FA and MD values of the 100 tensors as a measure of uncertainty in these parameters. We denote these as $\sigma^{\text{DL}}(\text{FA})$ and $\sigma^{\text{DL}}(\text{MD})$, respectively. To quantify subject/scan-level uncertainty, we compute the average over the entire brain volume.

\subsection{Data}

The bulk of the experiments in this work use the developing Human Connectome Project (dHCP) dataset \cite{bastiani2019}. We use all 88 measurements with $b=1000$ in each scan to estimate a ``ground truth" tensor estimation using the constrained weighted linear least squares (CWLLS) method \cite{koay2006unifying}. We select subsets of 30, 15, and 6 measurements from among the 88 measurements from each scan to train and evaluate our method. To select these subsets, we choose measurements that are closest to the optimized gradient directions proposed in prior works \cite{jones1999optimal,skare2000condition}. The dHCP dataset includes scans of newborns with gestational ages of 29-46 weeks. We use 100 scans with gestational ages of 40-46 weeks to train our model. We then test the trained model on 40 scans from the same age range. We also test the trained model on 48 scans with gestational ages of 29-36 weeks. Furthermore, we test the model on 25 scans from the Pediatric Imaging, Neurocognition, and Genetics (PING) dataset \cite{jernigan2016pediatric}. The subjects in the PING dataset are children and adolescents aged 9-20 years.

\begin{figure}[!htb]
  \centering
  \centerline{\includegraphics[width=9cm]{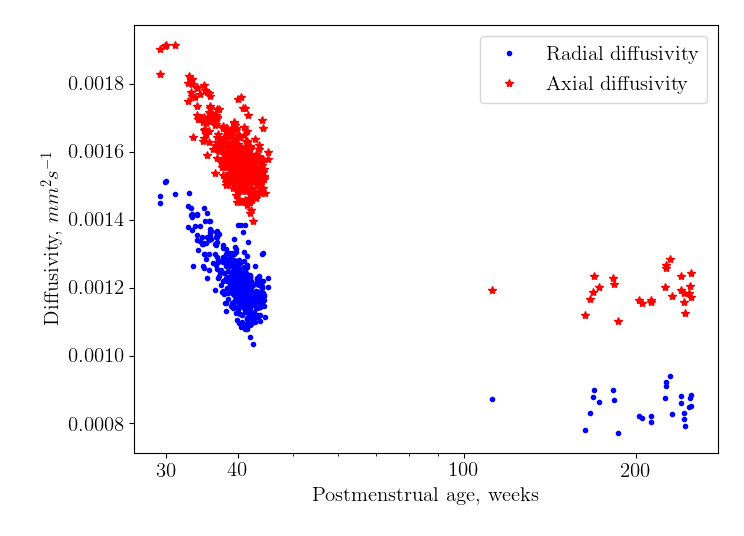}}
\caption{\footnotesize{Radial and axial diffusivity as a function of postmenstrual age for subjects in the dHCP (29-46 weeks) and PING datasets (9-20 years).}}
\label{fig:domain_shift}
\end{figure}

We use the scans of the younger subjects in the dHCP dataset and the scans from the PING dataset to assess possible changes in the performance of the DL model due to a domain shift. We investigate whether the estimation uncertainty is useful in detecting any changes in model performance. As shown in Figure \ref{fig:domain_shift}, tissue micro-structure changes dramatically as the brain matures. These changes are well documented in the published literature \cite{berman2005quantitative}. A machine learning model trained on a certain age range may not generalize well to another age range. This is a typical example of domain shift, also sometimes referred to as out-of-distribution data. A machine learning method can only be expected to work well on data that come from the distribution of the training data; test data samples that come from an entirely different distribution should be detected and flagged as out-of-distribution. In most applications this is a challenging problem as the data distribution is high-dimensional and complex. Different methods have been proposed to detect out-of-distribution data in deep learning. Some of the proposed methods are based on estimation uncertainty, where the hope is that data samples that are far from the training distribution result in high model uncertainty \cite{devries2018learning,hendrycks2016}. We examine the effectiveness of our proposed uncertainty estimations for this purpose.

\subsection{Evaluation criteria}

We quantify estimation error in terms of error in orientation of the major eigenvector, FA, and MD. Error is computed as the difference between the estimated value and ground truth.

We use two measures to quantify and compare uncertainty calibration. The first is similar to Expected Normalized Calibration Error (ENCE) \cite{levi2019evaluating}. Specifically, we compute RMV and RMSE using Eq. \ref{eq:RMV_RMSE}. Calibration can be assessed visually by plotting $\text{RMSE}$ versus $\text{RMV}$, where a perfect calibration will correspond to the identity function. For a quantitative assessment, we compute ENCE as:

\begin{equation}  \label{eq:ENCE}
\footnotesize
\text{ENCE}= \frac{1}{N} \sum_{j=1}^N \frac{ |B_j| . |\text{RMV}(j)- \text{RMSE}(j)|}{\text{RMV}(j)}.
\end{equation}

Compared with the definition of ENCE in \cite{levi2019evaluating}, we have included $|B_j|$ in the numerator to give larger weights to the bins with more data points. We think ENCE is the best measure of uncertainty calibration that we have studied in the literature. Yet, ENCE on its own is not adequate because a forecaster that predicts a single uncertainty value equal to the mean error over the entire population will have an ENCE of zero. Authors of \cite{levi2019evaluating} suggest also computing the coefficient of variation of $\sigma$.

Second, we propose a new measure of the quality of uncertainty estimations based on PICP and MPIW (Eq. \ref{eq:PICP_MPIW}). Prior works have analyzed these measures for a fixed confidence interval. Here, we propose a method of visualizing PICP and MPIW for a range of confidence intervals and an aggregate measure of confidence calibration. Consider a set $\{ (x_i, y_i ) \}_{i=1}^n$ of test samples $x_i$ (e.g., image voxels) and the corresponding ground truth parameter values $y_i$. Given $x_i$, a regression model predicts $\hat{y}_i$ and estimation uncertainty $\sigma_i$. Assuming symmetric uncertainty, the lower and upper bounds of the prediction interval can be constructed as ${y_L}_i(\beta)= \hat{y}_i - \beta \sigma_i$ and ${y_U}_i(\beta)= \hat{y}_i + \beta \sigma_i$. High-quality prediction intervals should 1) be narrow and 2) capture most of the observations \cite{pearce2018high,khosravi2010lower}. Prior works have computed PICP and MPIW for a single value of $\beta=1$. Here, we consider a range of $\beta \in [0, \beta_{\text{max}}]$:

\begin{equation}  \label{eq:PICP_beta}
\footnotesize
\begin{split}
\text{PICP}(\beta) & = \frac{1}{n} \sum_{i=1}^n \mathbb{I} \{ {y_L}_i(\beta) \leq y_i \leq {y_U}_i(\beta)   \} \\
\text{MPIW}(\beta) & = \frac{1}{n} \sum_{i=1}^n {y_U}_i(\beta) - {y_L}_i(\beta)
\end{split}
\end{equation}

$\text{MPIW}(\beta)$ is an increasing function of $\beta$. We choose $\beta_{\text{max}}$ such that $\text{MPIW}(\beta)$ reaches a pre-determined value that depends on the parameter of interest. For example for FA, which takes on values in [0, 1], typical estimation errors are in the range [0,0.20]. In this case, one may choose $\beta_{\text{max}}$ such that $\text{MPIW}(\beta_{\text{max}})=0.20$. We then plot $\text{PICP}(\beta)$ as a function of $\text{MPIW}(\beta)/ \max (\text{MPIW}(\beta))$. $\text{PICP}(\beta)$ can reach a maximum value of 1. Based on the high-quality principle \cite{pearce2018high}, in the ideal situation $\text{PICP}(\beta)$ should approach 1 for small values of $\text{MPIW}(\beta)$. Example plots are presented below (Figures \ref{fig:Recalibration_in_action} and \ref{fig:Recalibration_FA_MD}). These plots are reminiscent of the receiver operating characteristic curves in the field of signal detection. Following this analogy, we propose to compute the area under this curve as a measure of the quality of predicted estimation uncertainty. We denote this area with AUCC, for area under the calibration curve. AUCC takes on values between 0 and 1, with larger values indicating better uncertainty estimates. We think AUCC is a more appropriate measure of uncertainty calibration than those proposed in prior works including ENCE. Unlike ENCE, it is not prone to assigning high scores to degenerate uncertainty estimations. Moreover, it does take into account all measurements, allowing for the fact that different measurements are expected to have different uncertainty estimates. To achieve a higher AUCC, the model should predict smaller uncertainty values (tighter prediction intervals) for more accurate predictions and larger uncertainty values (wider prediction intervals) for less accurate predictions, which matches the definition of high-quality prediction intervals. AUCC is invariant to the scale of predicted uncertainties, which can be a positive or a negative quality.

\subsection{Post-hoc re-calibration}

To improve the calibration of computed uncertainties, we use post-hoc calibration using a held-out calibration dataset $(x_{\text{cal}}, y_{\text{cal}}) \in \mathcal{D}_{\text{cal}}$. We compute the estimation error and estimation uncertainty for $\mathcal{D}_{\text{cal}}$. We then divide the range of uncertainties into non-overlapping bins ${ \{ B_j \} }_{j=1}^m$ and for each bin compute the root mean variance (RMV) and RMSE as:

\begin{equation}  \label{eq:RMV_RMSE}
\footnotesize
\text{RMV}(j)= \sqrt{\frac{1}{|B_j|} \sum_{i \in B_j} \sigma_i^2 } , 
\text{RMSE}(j)= \sqrt{\frac{1}{|B_j|} \sum_{i \in B_j} (y_i - \hat{y}_i)^2 }
\end{equation}

Here we use variance ($\sigma^2$) and (the square of) uncertainty interchangeably. Definitions above are the same as those in \cite{levi2019evaluating}. According to the definition in Eq. \ref{eq:Levi}, perfect calibration is achieved when $\text{RMV}(j)= \text{RMSE}(j) \; \forall j$. Since RMSE is usually a non-decreasing function of RMV, we fit an isotonic regression function $R$ to the set of points $\big( \text{RMV}(j), \text{RMSE}(j) \big)$. On the test data, we apply $R$ on $\sigma^2$ to compute calibrated uncertainty estimates $\sigma_{\text{cal.}}^2= R \big( \sigma^2 \big)$.

\begin{figure*}[!htb]
  \centering
  \centerline{\includegraphics[width=18cm]{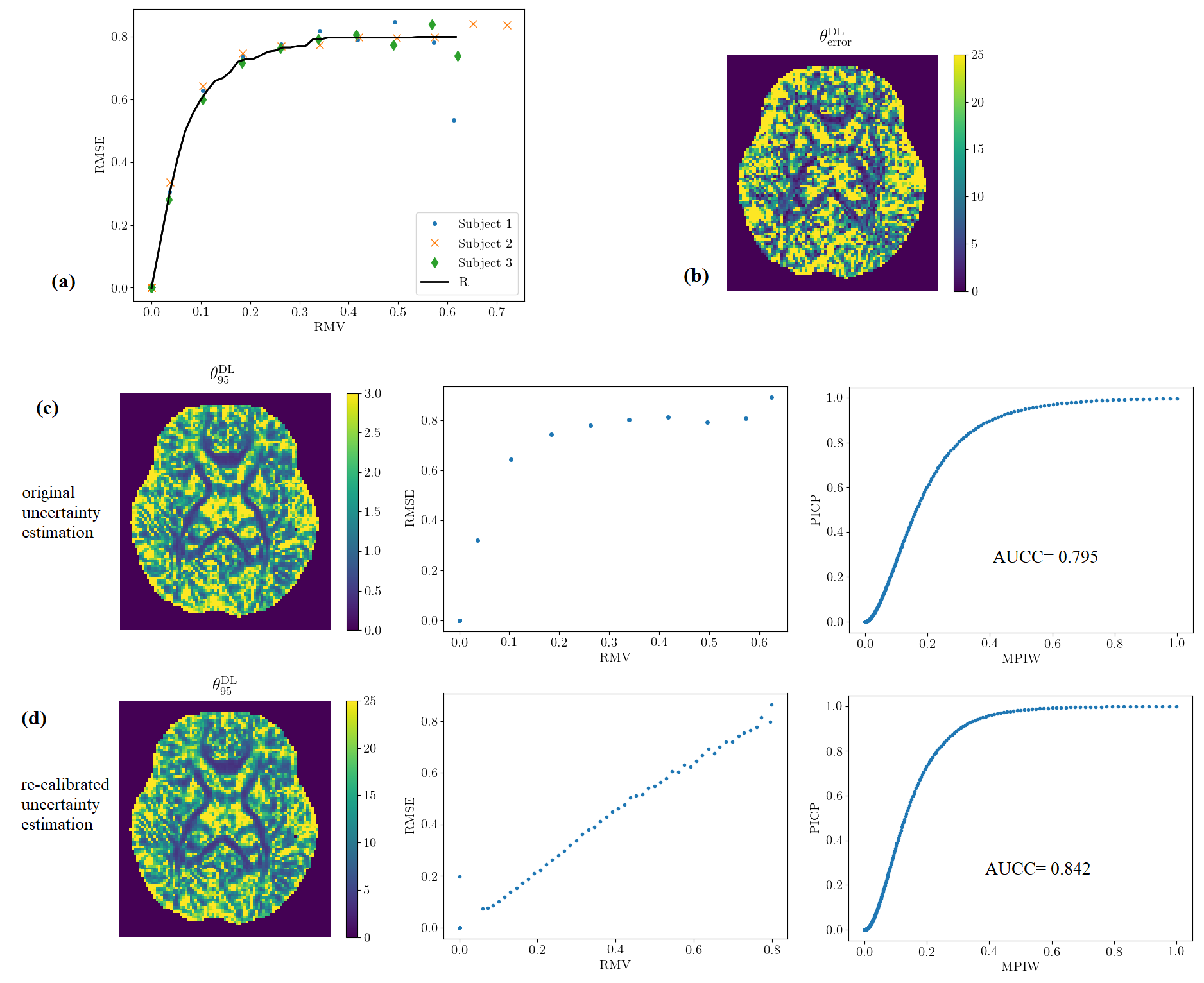}}
\caption{\footnotesize{The proposed post-hoc re-calibration method in action. This example shows the DL epistemic uncertainty for orientation of the major eigenvector on a dHCP test scan. (a) Plots of RMSE versus RMV for three training scans. The solid curve shows the isotonic regression function fitted to the data points for these three scans. This function is then used to re-calibrate the uncertainty estimations on a test scan. (b) Estimation error for the test scan. (c) Original (uncalibrated) uncertainty map for this subject and the corresponding calibration curves. (d) Uncertainty map after post-hoc re-calibration.}}
\label{fig:Recalibration_in_action}
\end{figure*}

Figure \ref{fig:Recalibration_in_action} demonstrates post-hoc re-calibration in action. The method is based on the observation that the uncertainty estimations on different scans are mis-calibrated in a similar way, as shown in the example plots in Figure \ref{fig:Recalibration_in_action}(a). Therefore, one can learn a function $R$ on a calibration set and use $R$ to re-calibrate the uncertainty estimations on the test data. As shown in the example in Figure \ref{fig:Recalibration_in_action}, unlike scaling techniques, this re-calibration method does not merely scale the uncertainty values. Rather, it also improves AUCC. Note that scaling the uncertainty values will not change the plot of PCIP versus MPIW. The example shown in Figure \ref{fig:Recalibration_in_action} is for orientation of the major eigenvector. We apply the same re-calibration approach to the epistemic uncertainty estimations for FA and MD.

\subsection{Compared methods}

We compare the DL framework described above with a standard set of methods. Specifically, we use CWLLS to estimate the tensor and the wild bootstrap (WBS) method \cite{whitcher2008using} to compute the estimation uncertainty. As suggested in \cite{zhu2008optimized}, we use 1000 iterations of the WBS. From the 1000 WBS-estimated tensors, we compute the estimation uncertainty for the orientation of the major eigenvector, FA, and MD in the same way as described above for the DL method. We denote these uncertainties with $\theta_{95}^{\text{WBS}}$, $\sigma^{\text{WBS}}(\text{FA})$ and $\sigma^{\text{WBS}}(\text{MD})$. For a fair comparison, we apply the same post-hoc re-calibration described above to also re-calibrate the WBS-estimated uncertainties.

\subsection{Implementation and training}

We implemented the DL method in TensorFlow. We trained and evaluated the model on a Linux computer with an NVIDIA GeForce GTX 1080 GPU.  We optimized the model weights using Adam \cite{kingma2014} with a batch size of 10 and initial learning rate of $10^{-4}$, which was reduced by half after every ten epochs if the validation loss did not decrease. Validation data consisted of 15 scans selected from among the 100 training scans. Training was stopped when the validation loss did not decrease in two consecutive evaluations. We ran CWLLS+WBS on CPU using the CWLLS implementation in DIPY \cite{garyfallidis2014dipy}.

\section{Results and Discussion}

\subsection{Estimation accuracy}

\begin{figure*}[!htb]
  \centering
  \centerline{\includegraphics[width=18cm]{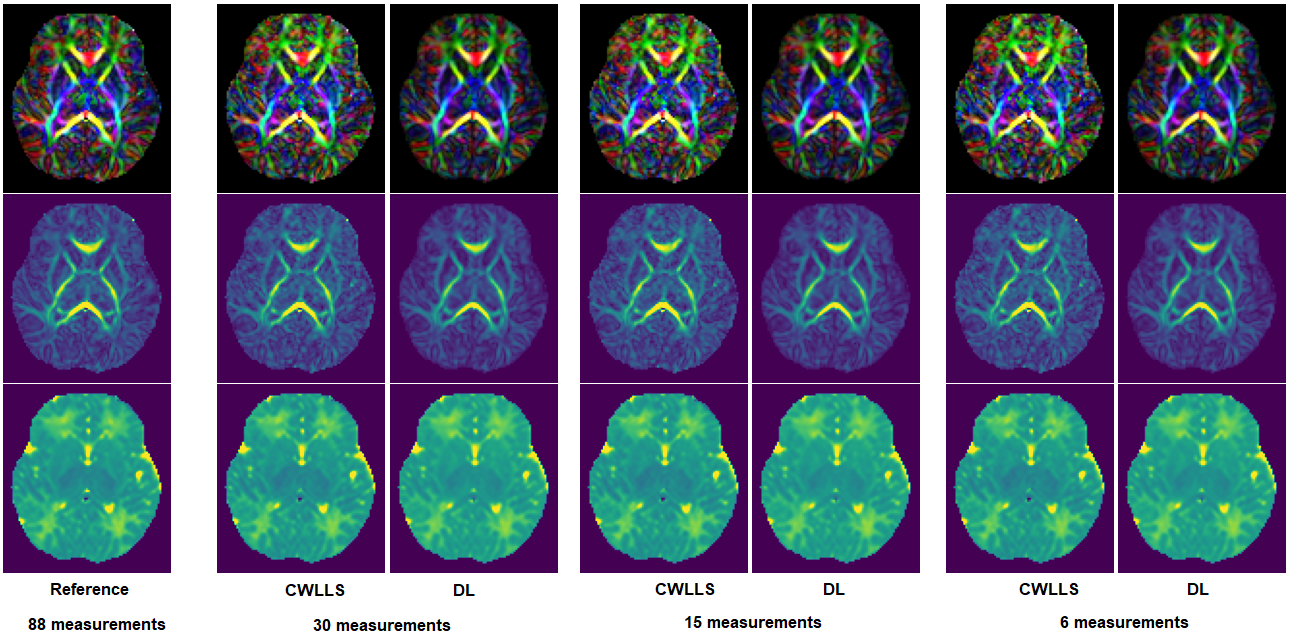}}
\caption{\footnotesize{Color FA, FA, and MD images estimated with CWLLS and DL methods on a test scan from the dHCP dataset.}}
\label{fig:accuracy_figure}
\end{figure*}

Table \ref{table:accuracy_table} shows tensor estimation errors in terms of FA, MD, and orientation of the major eigenvector. Figure \ref{fig:accuracy_figure} shows example color FA, FA, and MD images. Compared with CWLLS, the estimation accuracy of the DL method is much less affected by reducing the number of measurements from 30 to 6. Compared with the ground truth (estimated by CWLLS from 88 measurements) the DL method is more accurate in terms of FA and orientation of the major eigenvector when using 15 or 6 measurements for estimation. Given that the ground truth is also estimated with CWLLS, it may even have a bias against the DL method in this comparison.

\begin{table*}[!htb]
\centering
\scriptsize
 \caption{\footnotesize{Estimation error for CWLLS and the DL method in terms of the error in FA, MD, and orientation of the major eigenvector.}}
    \label{table:accuracy_table}
\begin{tabular}{ C{3.0cm}  L{1.2cm} C{2.5cm} C{4.5cm} C{2.5cm} }
\thickhline
number of measurements & Method & error(FA) & error(MD) $\times 1000$, $mm^2s^{-1}$  & error($\theta$), degrees  \\ \thickhline
\multirow{2}{*}{30}  & CWLLS & $0.038 \pm 0.003$ & $0.042 \pm 0.004$ & $17.8 \pm 1.53$   \\
& DL &                         $0.036 \pm 0.004$ & $0.045 \pm 0.004$ & $16.4 \pm 1.54$   \\
\hline
\multirow{2}{*}{15}  & CWLLS & $0.053 \pm 0.007$ & $0.047 \pm 0.005$ & $23.6 \pm 1.92$   \\
& DL                         & $0.041 \pm 0.004$ & $0.048 \pm 0.004$ & $19.0 \pm 1.56$   \\
\hline
\multirow{2}{*}{6}  & CWLLS & $0.111 \pm 0.014$ & $0.079 \pm 0.010$ & $33.4 \pm 1.77$   \\
& DL                         & $0.045 \pm 0.004$ & $0.056 \pm 0.006$ & $21.5 \pm 1.68$   \\
\thickhline
\end{tabular}
\end{table*}

\subsection{Estimation uncertainty}

\begin{figure*}[!htb]
  \centering
  \centerline{\includegraphics[width=18cm]{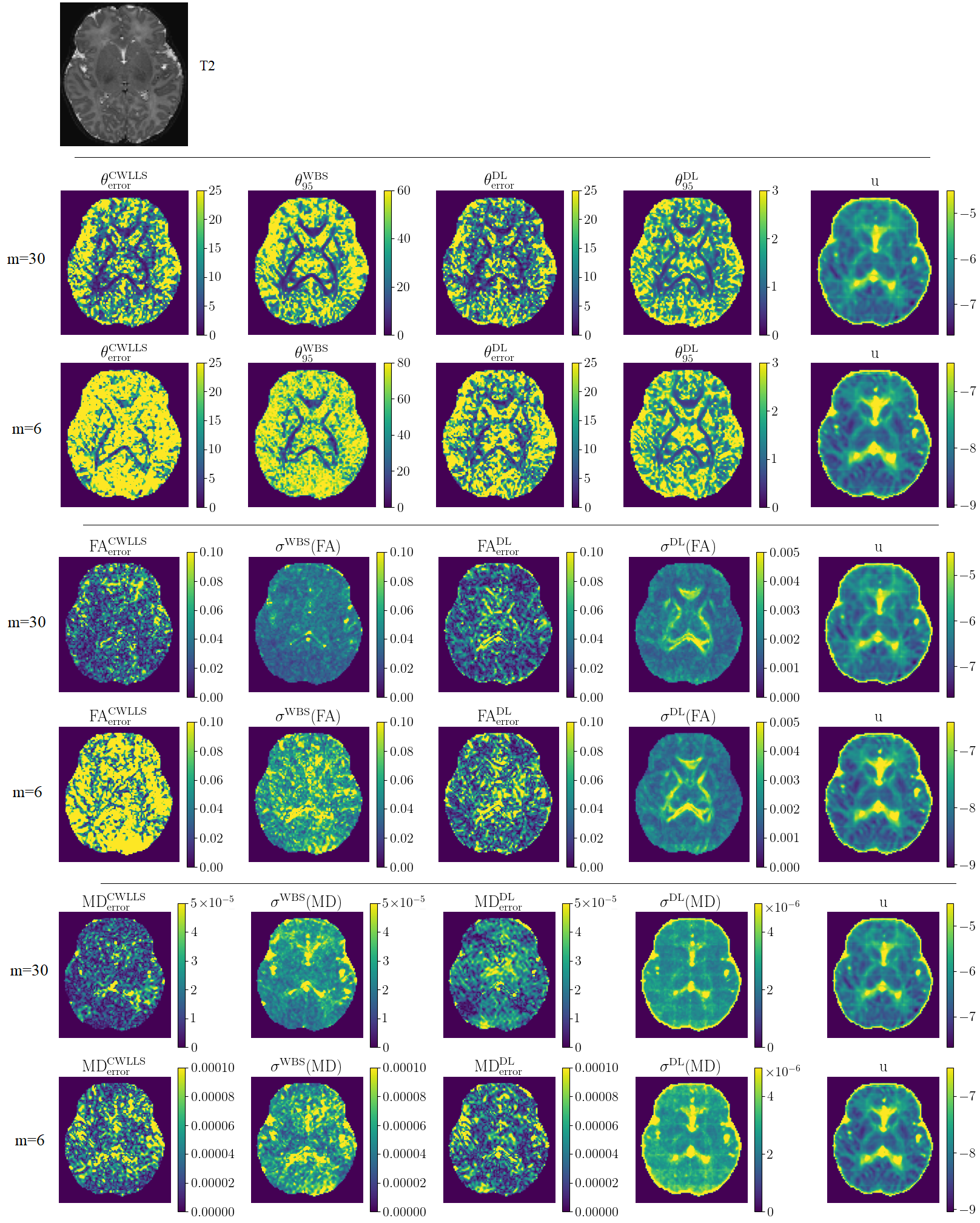}}
\caption{\footnotesize{Estimation error and uncertainty computed with CWLLS+WBS and the propsoed DL method. $m$ is the number of measurements.}}
\label{fig:raw_uncertainty_figure}
\end{figure*}

Figure \ref{fig:raw_uncertainty_figure} shows the estimation error and estimation uncertainty for orientation of the major eigenvector, FA, and MD for a test subject. It shows the results for the proposed DL method as well as for CWLLS+WBS method for estimation with 30 and 6 diffusion-weighted measurements. For the DL method, we have shown the epistemic uncertainty computed using Monte Carlo dropout as well as the aleatoric uncertainty, $u$. The latter is the same for all three parameters.

The first observation from Figure \ref{fig:raw_uncertainty_figure} is that the scales of uncertainty values are very different than the scales of the corresponding estimation errors for most parameters and estimation methods. This can be fixed by post-hoc uncertainty re-calibration method, as we show below. For orientation of the major eigenvector, both WBS-estimated uncertainties and epistemic DL uncertainties show low values for the location of major white matter tracts and high uncertainty for areas of lower FA. Overall, we observe the expected correlation between the estimation uncertainty and estimation error. For FA, the epistemic DL uncertainty is high for areas of higher FA, which is the opposite of the uncertainty pattern for orientation of the major eigenvector. The FA estimation uncertainty computed with WBS does not show a clear pattern. For MD, both epistemic DL uncertainty and WBS-estimated uncertainty are higher for CSF voxels. The aleatoric uncertainty, $u$, estimated with the DL method is highest for CSF voxels, followed by white matter voxels.

\begin{figure*}[!htb]
  \centering
  \centerline{\includegraphics[width=18cm]{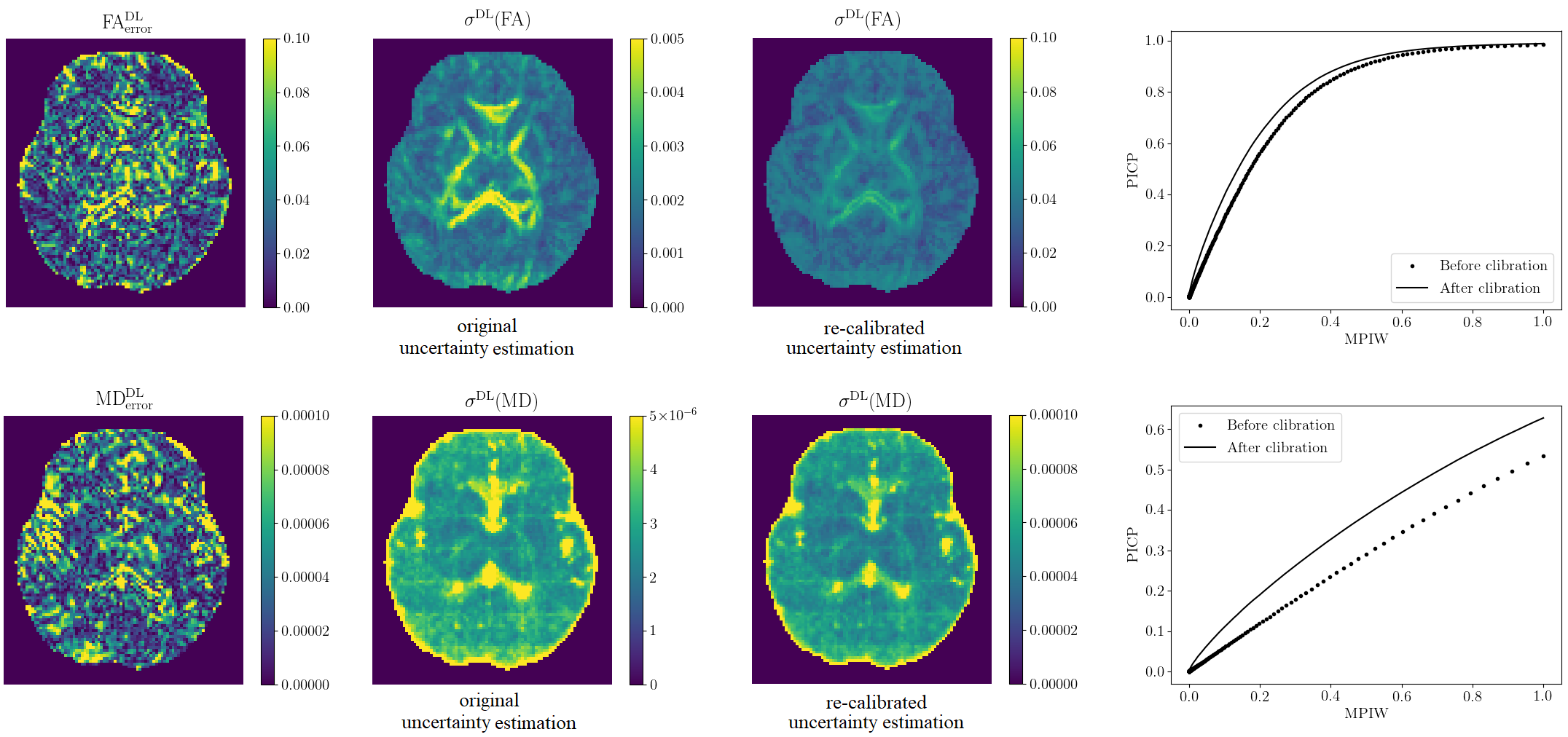}}
\caption{\footnotesize{Example post-hoc re-calibration results for DL epistemic uncertainties for FA (top) and MD (bottom).}}
\label{fig:Recalibration_FA_MD}
\end{figure*}

\begin{figure*}[!htb]
  \centering
  \centerline{\includegraphics[width=18cm]{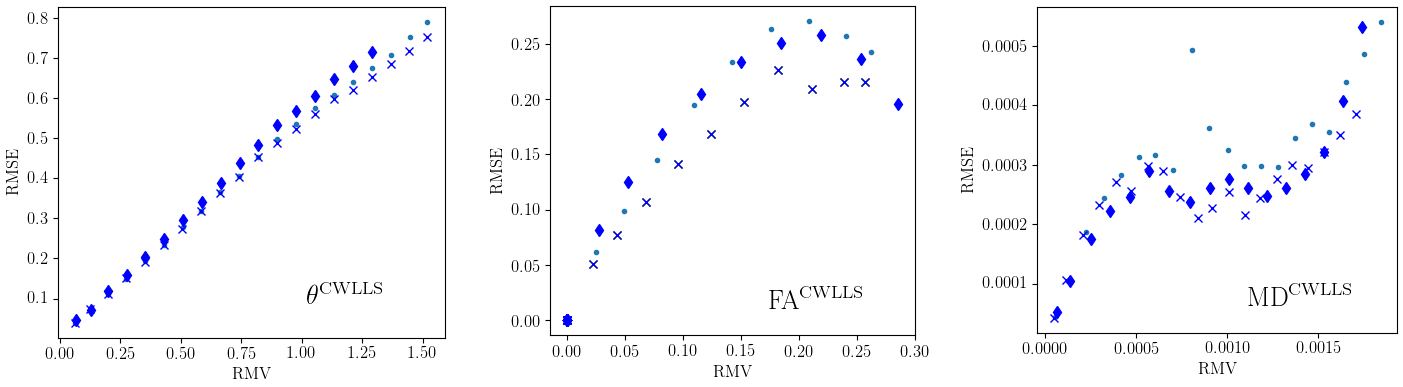}}
\caption{\footnotesize{Example plots of RMSE vs. RMV estimated with WBS. Each figure shows the plots for three independent scans from the dHCP dataset.}}
\label{fig:Calibration_WBS}
\end{figure*}

Before evaluating the uncertainty estimations in terms of ENCE and AUCC, we apply post-hoc re-calibration. The example shown in Figure \ref{fig:Recalibration_in_action} above is for orientation of the major eigenvector. Example re-calibration results for FA and MD estimation uncertainties are shown in Figure \ref{fig:Recalibration_FA_MD}. We apply the same re-calibration method to uncertainties estimated with CWLLS+WBS. As shown in the example plots in Figure \ref{fig:Calibration_WBS}, the plots of RMV versus RMSE are largely similar across subjects. Therefore, post-hoc re-calibration works for CWLLS+WBS as well. In fact, as shown in Figure  \ref{fig:Calibration_WBS}(a), the plots for orientation of the major eigenvector are surprisingly close to linear. Hence, a scaling would be adequate in this case.

Table \ref{table:calibration_table} shows values of ENCE and AUCC for the proposed DL method and CWLLS+WBS. These values are computed after re-calibration. Overall, CWLLS+WBS has computed better-calibrated estimation uncertainties than the DL method with 30 measurements, whereas when using six measurements the uncertainties computed by the DL method are better calibrated. In terms of AUCC, the uncertainties for orientation of the major eigenvector and FA are much better calibrated than for MD. ENCE values depend on the scale of the parameter of interest and hence they are naturally much smaller for MD. Overall, the computed uncertainties for MD seem to be poorly calibrated. On the other hand, MD is known to be easier to estimate \cite{jones2009gaussian}, which is also born out by our results (Table \ref{table:accuracy_table}) that show little increase in estimation error when the number of measurements is reduced to six. It is also encouraging to note that the differences between methods in terms of ENCE are largely consistent with the differences in terms of AUCC, although they are based on two different notions of uncertainty/confidence calibration. Note that lower ENCE and higher AUCC indicate better calibration.

\begin{table*}[!htb]
  \centering
\scriptsize
 \caption{\footnotesize{Quantitative assessment of uncertainty estimations computed with the proposed DL method and CWLLS+WBS method. $m$ is the number of measurements.}}
    \label{table:calibration_table}
\begin{tabular}{ L{2.0cm} L{1.0cm}  L{2.2cm} C{2.2cm} C{2.2cm} C{2.2cm} }
\thickhline
& $m$ & Method & FA & MD & $\theta$   \\ \thickhline
\multirow{4}{*}{ENCE $\times 1000$} & \multirow{2}{*}{30}  & CWLLS+WBS & $0.252 \pm 0.027$ & $0.0033 \pm 0.0021$ & $0.225 \pm 0.041$   \\
&                                               & DL & $0.281 \pm 0.031$ & $0.0034 \pm 0.0018$ & $0.243 \pm 0.035$   \\
\cline{2-6}
& \multirow{2}{*}{6}                         & CWLLS+WBS & $0.721 \pm 0.059$ & $0.0063 \pm 0.0031$ & $0.276 \pm 0.055$   \\
&                                               & DL & $0.680 \pm 0.045$ & $0.0052 \pm 0.0028$ & $0.260 \pm 0.058$   \\
\thickhline
\multirow{4}{*}{AUCC} & \multirow{2}{*}{30}  & CWLLS+WBS & $0.834 \pm 0.110$ & $0.444 \pm 0.104$ & $0.874 \pm 0.081$   \\
&                                               & DL & $0.810 \pm 0.094$ & $0.429 \pm 0.108$ & $0.853 \pm 0.075$   \\
\cline{2-6}
& \multirow{2}{*}{6}                         & CWLLS+WBS & $0.732 \pm 0.125$ & $0.246 \pm 0.128$ & $0.771 \pm 0.062$   \\
&                                               & DL & $0.783 \pm 0.140$ & $0.328 \pm 0.136$ & $0.800 \pm 0.079$   \\
\thickhline
\end{tabular}
\end{table*}

\subsection{Practical utility of estimation uncertainty}

In this section, we demonstrate the practical utility of uncertainty estimation, with a focus on the DL method. Estimation uncertainty should ideally be correlated with or indicative of estimation error. In presenting our experimental results above, we have shown estimation uncertainty side by side with estimation error, which we have computed using a ``ground truth". However, in practice we usually do not know the ground truth. Estimation uncertainty values should inform us of large or unusual estimation errors. Here, using several examples, we show that the uncertainty maps computed with the proposed methods can serve this purpose.

\begin{figure*}[!htb]
  \centering
  \centerline{\includegraphics[width=18cm]{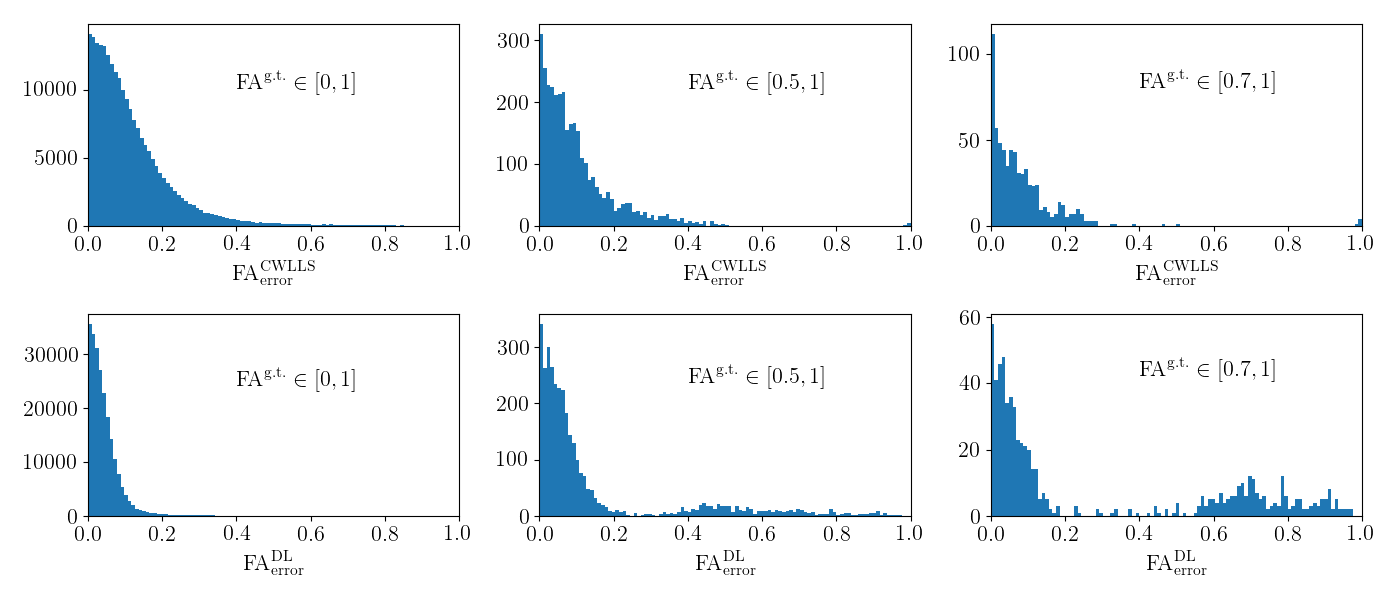}}
\caption{\footnotesize{Histograms of FA estimation error for CWLLS (top row) and the DL method (bottom row). These are the results on a test scan from the dHCP dataset. We observed very similar results on all test scans. $\text{FA}^{\text{g.t.}}$ denotes the ground-truth FA.}}
\label{fig:FA_error}
\end{figure*}

As we have pointed out above, the DL method shows higher FA uncertainties for higher FA values (cf. Figure  \ref{fig:raw_uncertainty_figure}). In order to see whether these high uncertainty values highlight large estimation errors, in Figure \ref{fig:FA_error} we have shown histograms of FA estimation errors for CWLLS and DL methods for different ranges of ground truth FA values. As can be seen in this figure, although the DL method achieves significantly smaller mean estimation error than CWLLS over the whole brain, its error is actually \emph{larger} for high FA values. This estimation bias is not specific to our network architecture. We have observed this bias with a few other architectures, including an architecture that has recently been propsoed for DTI estimation in \cite{li2021superdti}. Depending on the application, this can be a serious bias. For example, FA values on major white matter tracts are often used as biomarkers of white matter development and degeneration. This example shows the potential of uncertainty maps to highlight such biases.

\begin{figure*}[!htb]
  \centering
  \centerline{\includegraphics[width=18cm]{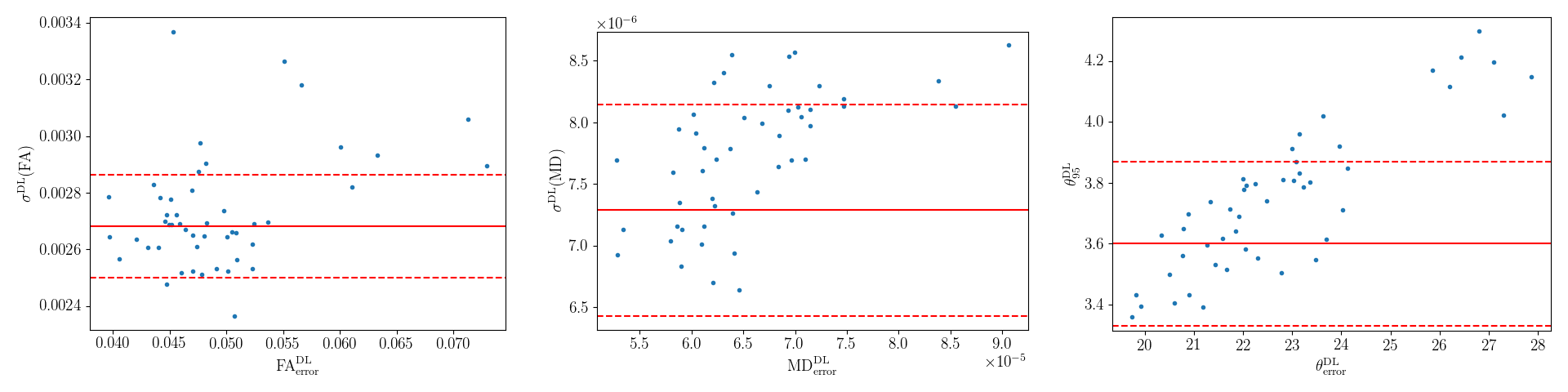}}
\caption{\footnotesize{Plots of estimation uncertainty versus estimation error for the proposed DL method on test dHCP subjects with gestational ages between 29 and 36 weeks. The DL model has been trained on subjects with gestational ages between 40 and 46 weeks. In each plot, the solid horizontal line shows the mean estimation uncertainty on training scans; the dashed horizontal lines show the mean $\pm \, 2.0 $ standard deviations.}}
\label{fig:young_dHCP_correlations}
\end{figure*}

To assess the impact of domain shift, we first show the performance of the trained model on the younger subjects in the dHCP dataset. Table \ref{table:accuracy_young_table} shows the estimation error of the trained model on those subjects. A comparison with Table \ref{table:accuracy_table}, where test subjects have the same age range as the training subjects, shows that estimation errors in Table \ref{table:accuracy_young_table} are slightly larger than those in Table \ref{table:accuracy_table}. Figure \ref{fig:young_dHCP_correlations} shows plots of epistemic uncertainty versus estimation error for these younger subjects. There is a strong positive correlation between uncertainty and error. The Pearson correlation coefficients between error and uncertainty are 0.376, 0.591, and 0.868 for FA, MD, and $\theta$, respectively. Test subjects with higher estimation errors can be detected based on their higher uncertainties, especially in terms of $\theta$. This example does not represent a significant domain shift as the trained model still performs well and is more accurate than CWLLS. Nonetheless, the results are encouraging as they show that the uncertainties can mark small increases in estimation error due to a slight domain shift.

\begin{table*}[!htb]
\centering
\scriptsize
 \caption{\footnotesize{Estimation errors of CWLLS and the DL method on test dHCP subjects with gestational ages between 29 and 36 weeks. The DL model has been trained on subjects with gestational ages between 40 and 46 weeks.}}
    \label{table:accuracy_young_table}
\begin{tabular}{ C{3.0cm}  L{1.2cm} C{2.5cm} C{4.5cm} C{2.5cm} }
\thickhline
number of measurements & Method & error(FA) & error(MD) $\times 1000$, $mm^2s^{-1}$  & error($\theta$), degrees  \\ \thickhline
\multirow{2}{*}{30}  & CWLLS & $0.037 \pm 0.003$ & $0.044 \pm 0.004$ & $18.1 \pm 1.56$   \\
& DL                         & $0.040 \pm 0.005$ & $0.044 \pm 0.005$ & $16.9 \pm 1.55$   \\
\hline
\multirow{2}{*}{15}  & CWLLS & $0.055 \pm 0.005$ & $0.045 \pm 0.006$ & $24.8 \pm 2.03$   \\
& DL                         & $0.054 \pm 0.007$ & $0.046 \pm 0.007$ & $20.1 \pm 1.74$   \\
\hline
\multirow{2}{*}{6}  & CWLLS & $0.107 \pm 0.011$ & $0.084 \pm 0.011$ & $33.6 \pm 1.80$   \\
& DL                        & $0.049 \pm 0.007$ & $0.066 \pm 0.008$ & $22.7 \pm 2.03$   \\
\thickhline
\end{tabular}
\end{table*}

\begin{figure*}[!htb]
  \centering
  \centerline{\includegraphics[width=18cm]{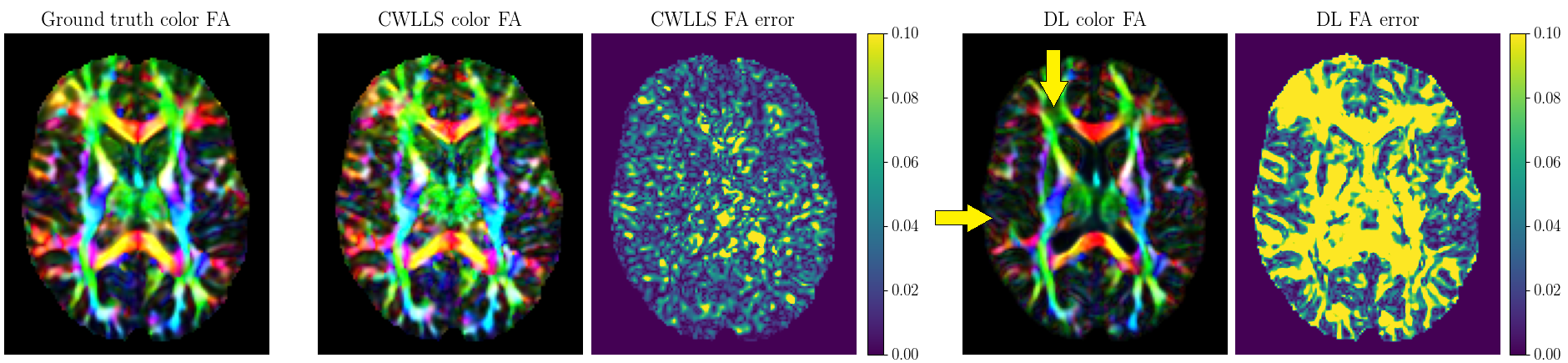}}
\caption{\footnotesize{Example results produced by CWLLS and the DL method (trained on the dHCP dataset) on a test scan from the PING dataset. Arrows on the color FA image predicted by the DL method point to the locations of some of the glaring errors. (Each scan in the PING dataset includes 30 diffusion-weighted measurements at $b=1000$. All 30 measurements are used to estimate the ground truth. CWLLS and DL method are applied on six measurements.)}}
\label{fig:ping_error}
\end{figure*}

\begin{figure*}[!htb]
  \centering
  \centerline{\includegraphics[width=18cm]{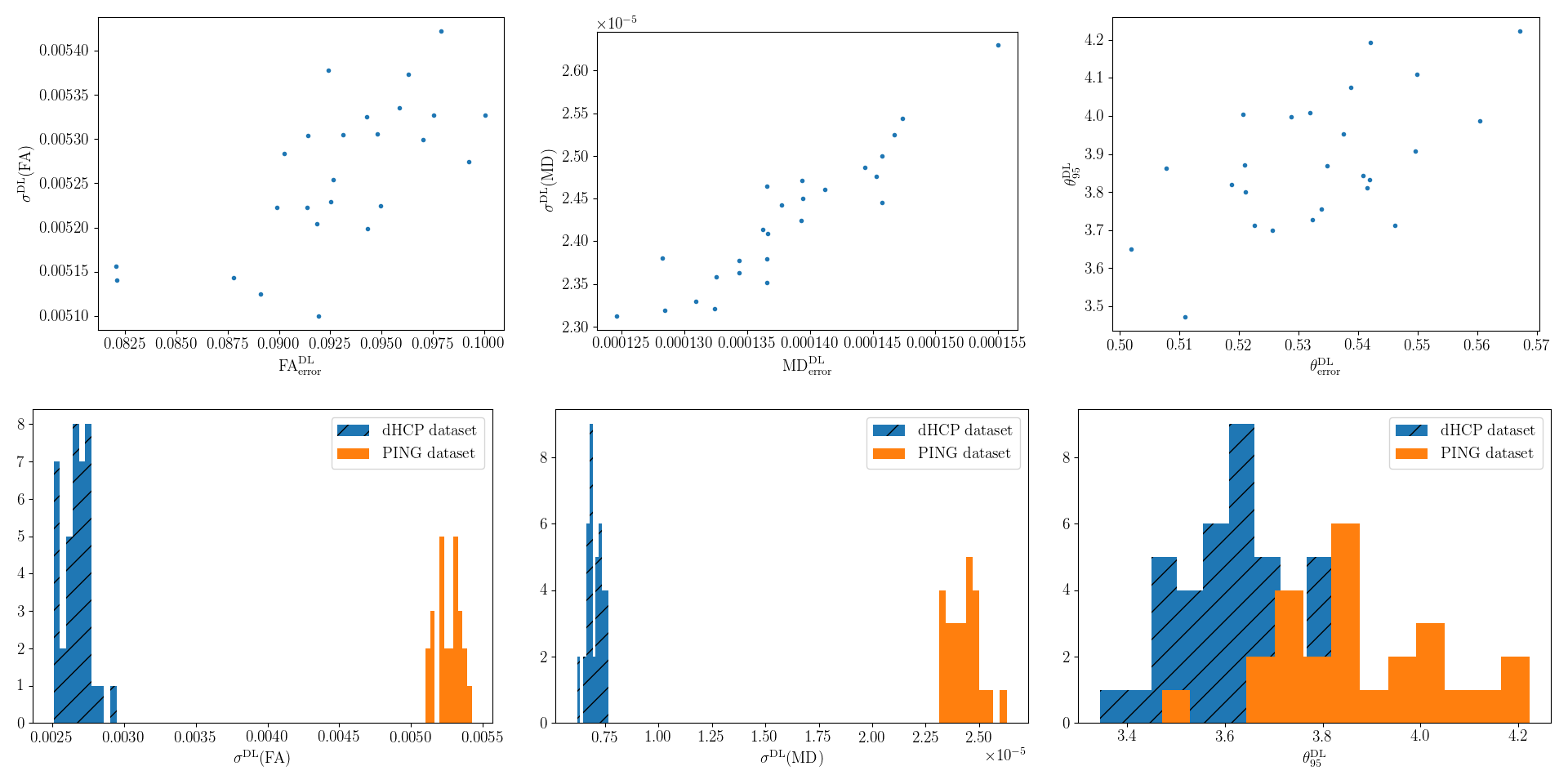}}
\caption{\footnotesize{Top: Plots of estimation uncertainty versus estimation error for the DL method on test scans from the PING dataset (children and adolescents). The DL model has been trained on the dHCP dataset (newborns). Bottom: histograms of the epistemic uncertainty values for the in-distribution (dHCP) data and out-of-distribution (PING) data. Especially in terms of the epistemic uncertainty of FA and MD, the out-of-distribution data are nicely separated.}}
\label{fig:ping_correlations}
\end{figure*}

On the PING dataset, on the other hand, the dHCP-trained model performed very poorly. It achieved mean estimation error for orientation of the major eigenvector, FA, and MD of $30.5^{\circ}$, 0.093, and $0.318 \time 10^{-3} mm^2s^{-1}$, respectively, which were much larger than those for the in-distribution test data (Table \ref{table:accuracy_table}). Example results are shown in Figure \ref{fig:ping_error}. The DL method can reconstruct the orientations of the major white matter tracts reasonably well (as observed on the color FA image), but it has large errors (marked with arrows) and produces highly inaccurate FA estimation. Figure \ref{fig:ping_correlations} shows plots of estimation uncertainty versus estimation error for the 25 test scans from the PING dataset. There is a strong correlation between uncertainty and error. The Pearson correlation coefficients between uncertainty and error are 0.677, 0.923, and 0.592 for FA, MD, and $\theta$, respectively. More importantly, the uncertainties for the scans in this dataset are significantly higher than for the training data, as can be seen in the histograms. The epistemic uncertainties for FA and MD perfectly separate these scans from the training scans. This encouraging observation shows that the epistemic uncertainties estimated by the DL method can be used to effectively detect out-of-distribution test data.

\begin{figure*}[!htb]
  \centering
  \centerline{\includegraphics[width=12cm]{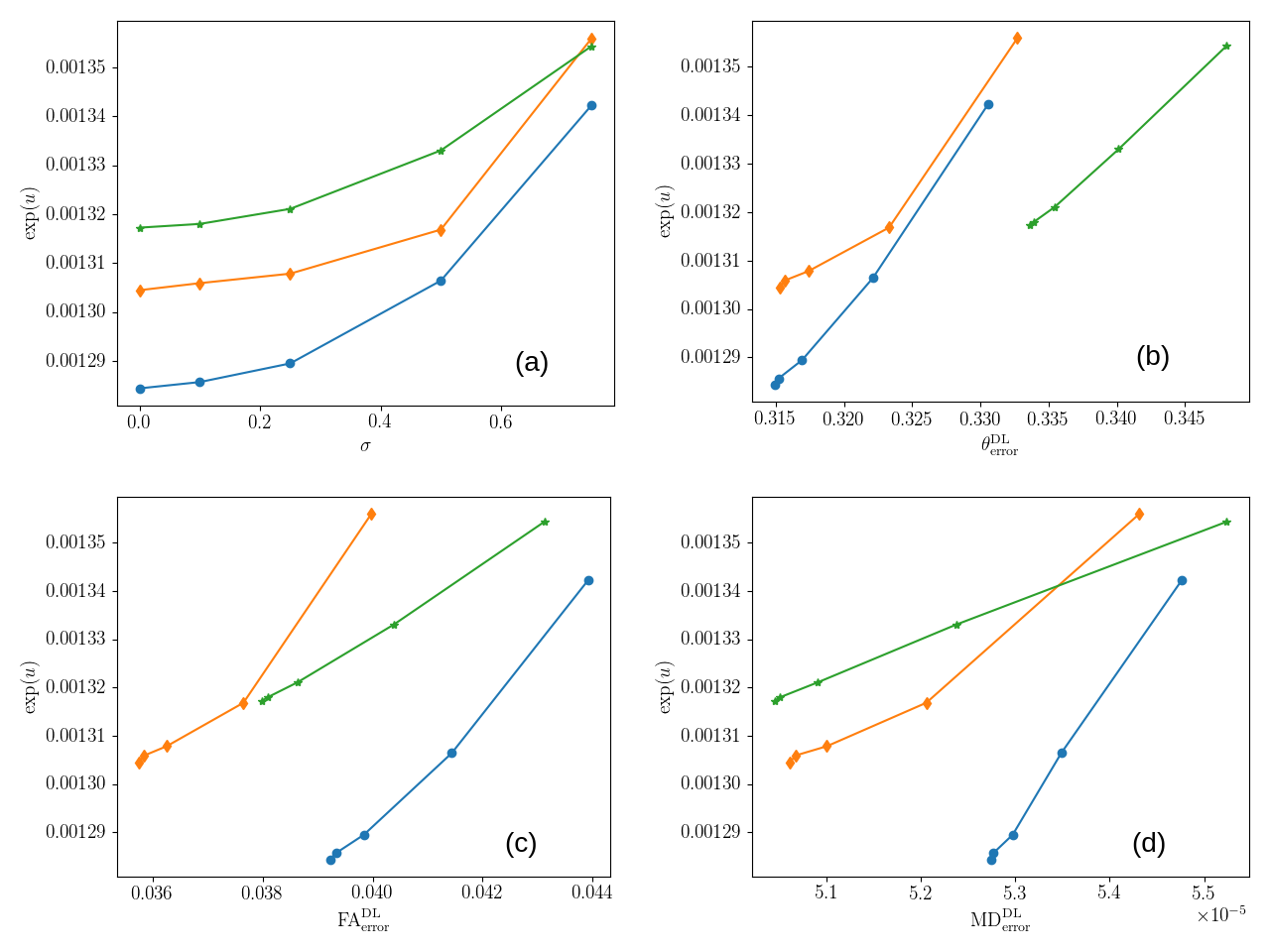}}
\caption{\footnotesize{Data-dependent uncertainty ($u$) as a function of noise standard deviation (a) and as a function of estimation error (b-d). In this experiment, Gaussian noise with varying standard deviation ($\sigma$) was added to scans from the dHCP dataset. In these plots, each of the three curves shows the result for a different subject.}}
\label{fig:noise_dependent_uncertainty}
\end{figure*}

In the experimental results presented above, we have used exclusively the epistemic uncertainty to detect large estimation errors and domain shift. The data-dependent uncertainty, $u$, on the other hand, is mainly intended to capture the data-dependent uncertainty and not the uncertainty in the model. To show that $u$ can capture data-dependent uncertainty, we performed a simulation experiment where we added different amounts of noise to the diffusion-weighted measurements in the dHCP scans. Specifically, we used Rician noise with signal to noise ratio (SNR) in the range [20,35] dB. Clearly, this simulated noise is added on top of the measurement noise. Figure \ref{fig:noise_dependent_uncertainty} shows the computed data-dependent uncertainty as a function of SNR for three test subjects. It also shows the relation between this uncertainty and the estimation error for orientation of major eigenvector, FA, and MD. There is a clear monotonic trend in all these plots. The results of this experiment show that the data-dependent uncertainty computed by the proposed DL model has the potential to predict measurement noise and the concomitant increase in estimation error.

\section{Conclusions and future directions}

It is likely that interest in DL-based methods for analyzing dMRI data will continue to grow. Safe and reliable deployment of these methods for clinical and research applications will require a proper understanding of their shortcomings, limits of generalizability, and failure modes. These topics have been largely neglected in prior studies. Our study shows that it is possible to estimate informative and practically useful uncertainties for DL-based dMRI parameter estimation methods. Our results show that these uncertainties can signal unusual or large errors, such as the large errors in estimating high FA values that we have observed in our experiments. They can also signal model failure due to domain shift. This study has only considered domain shift due to a change in the subjects' age. Many other factors related to the subjects or MRI signal acquisition may cause significant domain shift, which should be studied in future works. The role of alternative training strategies, such as adversarial training \cite{lee2017training,lakshminarayanan2017}, in improving the uncertainty calibration and out-of-distribution detection could also be investigated in future studies.

\section*{Acknowledgments}

This research was supported in part by the National Institute of Neurological Disorders and Stroke and Eunice Kennedy Shriver National Institute of Child Health and Human Development of the National Institutes of Health (NIH) under award numbers R01HD110772 and R01NS128281, by NIH grants R01EB031849, R01NS106030, R01EB032366, and S10OD0250111, and in part by the National Science Foundation (NSF) under award 2123061. The content of this publication is solely the responsibility of the authors and does not necessarily represent the official views of the NIH.

\bibliographystyle{ieeetr}
\bibliography{davoodreferences}

\end{document}